\documentclass[10pt,conference]{IEEEtran}
\usepackage{lipsum}
\usepackage{times,amsmath,amssymb}
\usepackage{graphicx}
\usepackage{xcolor}
\usepackage{algorithm2e}
\usepackage{algorithmicx}
\usepackage{setspace}
\usepackage[export]{adjustbox}
\usepackage{multirow}
\usepackage[noend]{algpseudocode}
\usepackage{array,multirow}
\usepackage{makecell}

\DeclareMathOperator*{\argmin}{arg\,min}

\let\mr=\multirow

\begin{document}
	
	\title{Performance Indicator in Multilinear Compressive Learning}
	\author{\IEEEauthorblockN{Dat Thanh Tran\IEEEauthorrefmark{1}, Moncef Gabbouj\IEEEauthorrefmark{1} \& Alexandros Iosifidis\IEEEauthorrefmark{2}}
		\IEEEauthorblockA{\IEEEauthorrefmark{1}Department of Computing Sciences, Tampere University, Finland\\
			\IEEEauthorrefmark{2}Department of Engineering, Aarhus University, Denmark\\
			Email:\{thanh.tran,moncef.gabbouj\}@tuni.fi, alexandros.iosifidis@eng.au.dk}\\
		
	}
	
	\maketitle
	
	\begin{abstract}
		Recently, the Multilinear Compressive Learning (MCL) framework was proposed to efficiently optimize the sensing and learning steps when working with multidimensional signals, i.e. tensors. In Compressive Learning in general, and in MCL in particular, the number of compressed measurements captured by a compressive sensing device characterizes the storage requirement or the bandwidth requirement for transmission. This number, however, does not completely characterize the learning performance of a MCL system. In this paper, we analyze the relationship between the input signal resolution, the number of compressed measurements and the learning performance of MCL. Our empirical analysis shows that the reconstruction error obtained at the initialization step of MCL strongly correlates with the learning performance, thus can act as a good indicator to efficiently characterize learning performances obtained from different sensor configurations without optimizing the entire system.  
	\end{abstract}
	
	\IEEEpeerreviewmaketitle
	
	\section{Introduction}
	
	Compressive Sensing (CS) \cite{candes2008introduction} is an efficient signal acquisition method that acquires the measurement of the signal by sampling and linearly interpolating the samples at the hardware level, i.e. by using CS devices. Particularly, let $\mathbf{y} \in \mathbb{R}^{I}$ be the discrete measurements of the input signal. Using a CS device, we obtain the compressed measurements $\mathbf{z}$ of the signal, instead of $\mathbf{y}$, with the compression step as follows:
	\begin{equation}\label{eq1}
	\mathbf{z} = \mathbf{\Phi} \mathbf{y}
	\end{equation}
	where $\mathbf{z} \in \mathbb{R}^{M}$ often has significantly lower dimension than $\mathbf{y}$, i.e., $M \ll I$. $\mathbf{\Phi} \in \mathbb{R}^{M \times I}$ is called the sensing operator. 
	
	This is different from the traditional approach where we obtain the discrete samples $\mathbf{y}$ from the signal acquisition device, and compression step is often conducted at the software level, \textit{being separate from the acquisition step}. Since signal compression is performed before signal registration during the sampling phase, CS devices require significantly lower temporary storage and bandwidth requirement. This paradigm is therefore prevalent in many applications that involve high-dimensional signals or critical computational requirements. 
	
	Although, in general, ideal sampling requires the signal to be sampled at higher rates than the Nyquist rate to ensure perfect reconstruction, in CS, the undersampled signal (due to $M \ll I$) can still be reconstructed almost perfectly if the sparsity assumption holds and the sensing operators possess certain properties \cite{candes2006stable, donoho2006compressed}. While the possibility to recover $\mathbf{y}$ from compressed measurements $\mathbf{z}$ is critical in some applications, like Magnetic Resonance Imaging (MRI) for expert diagnosis, there are other applications where the main goal is to detect certain patterns or to infer special properties from the acquired signal, rather than signal recovery. Thus, arises the idea of learning from compressed measurement. 
	
	Compressive Learning (CL) \cite{calderbank2012finding, davenport2007smashed, davenport2010signal, reboredo2013compressive} combines Compressive Sensing and Machine Learning into a single optimization problem which focuses on maximizing the learning performance, rather than performance on signal reconstruction. In the early works, the design of the sensing operator $\mathbf{\Phi}$ was decoupled from the construction of the learning model. Following the developments and wide adoption of stochastic optimization, recent works \cite{adler2016compressed, lohit2016direct, hollis2018compressed, zisselman2018compressed, tran2019multilinear, tran2020multilinear} have adopted an end-to-end learning paradigm that jointly optimizes the sensing operator and the inference model. 
	
	In order to work efficiently with multidimensional signals, Multilinear Compressive Learning (MCL) was recently proposed in \cite{tran2019multilinear}. MCL formulates sensing and feature synthesis based on multilinear algebra. Multilinear sensing and feature synthesis operators not only preserve the natural tensor format of the multidimensional signal but also require fewer computations and memory compared to other CL models which operate on vectorized signals. This makes MCL highly suitable for applications requiring the analysis of high-dimensional signals like images/videos on constrained computation/bandwidth platforms, such as drones and robots. 
	
	When building an MCL model to tackle a particular learning task, the configuration of the CS device plays an important role in the design process. Particularly, the choice of input resolution ($I$), i.e., the number of discrete samples initially captured by the device, and the size of compressed measurements ($M$) directly affects the learning performance. $I$ and $M$ characterize the computational complexity of CS device while $M$ alone characterizes the requirement for storage or transmission bandwidth. Given that a small increment or decrement of $I$ and/or $M$ only leads to a small changes in computational complexity which might be within the design requirements, extensive experimentation is needed to determine the a good combination of the dimensions $I$ and $M$ for the problem at hand.   
	
	In this work, by analyzing the performance under different combinations of $I$ and $M$, we seek to find a performance indicator of MCL models that can help us rapidly gauge different configurations of the CS component without the need of conducting the entire optimization process. Our empirical analysis reveals that the reconstruction error obtained at the initialization step of MCL models strongly correlates with the final learning performance, thus can act as a good performance indicator. 
	
	\section{Related Work}
	
	We are not aware of any work that aims to characterize the learning performance of a Compressive Learning system in terms of the sensing configurations or that investigates possible surrogate measures for its performance. Existing works only evaluate few configurations of the compressed measurement or the resolution of the input signal, while their experiments are not designed to isolate the effect of CS device configuration for studying its importance to the final learning performance. Remotely related to our work is the class of Neural Architecture Search (NAS) methods \cite{elsken2018neural, tran2019heterogeneous, tran2020progressive, kiranyaz2020operational} that estimates the performance of a candidate architecture by a surrogate model \cite{kandasamy2018neural, liu2018progressive} or by learning curve extrapolation \cite{domhan2015speeding, klein2016learning}. Instead of learning to predict the performance, we extensively evaluate several configurations of CS device on different learning problems and analyze the results to seek for a consistent performance indicator. 
	
	A Multilinear Compressive Learning (MCL) system \cite{tran2019multilinear} consists of three modules: the Compressive Sensing (CS) component, the Feature Synthesis (FS) component and the task-specific neural network $\mathfrak{N}$. 
	
	The Compressive Sensing (CS) component of MCL adopts multidimensional compressive sensing which is implemented via separable sensing operators, each of which operates on a mode of the input signal (tensor). Specifically, let us denote by $\mathcal{Y} \in \mathbb{R}^{I_1 \times \dots \times I_K}$ and $\mathcal{Z} \in \mathbb{R}^{M_1 \times \dots \times M_K}$ the discrete samples of the input tensor signal and the compressed measurements obtained from CS component, respectively. Here $I_1 \times \dots \times I_K$ denotes the resolution of sensors of the CS device. The CS component performs signal acquisition as follows:
	\begin{equation}\label{eq2}
	\mathcal{Z} = \mathcal{Y} \times_1 \mathbf{\Phi}_1 \times \dots \times_K \mathbf{\Phi}_K
	\end{equation} 
	where $\mathbf{\Phi}_k \in \mathbb{R}^{M_k \times I_k}$, $k=1, \dots, K$ denote the separable sensing operators and $\times_k$ denotes the mode-$k$ product. 
	
	Here we should note that the CS device performs discrete sampling (obtaining $\mathcal{Y}$) and compression simultaneously, producing the compressed measurements $\mathcal{Z}$ as the digital output, while $\mathcal{Y}$ is not registered digitally. $\mathcal{Y}$ can be considered as an intermediate state of the signal when acquired by a CS device. The dimension of $\mathcal{Y}$ represents the resolution at which the sensor inside a CS device performs discrete sampling. In addition, when the system is deployed, the CS component is implemented at the hardware level with configuration parameters $\mathbf{\Phi}_k$. That is, the CS component is a signal acquisition device and what we obtain from this device is the compressed version of the signal, i.e. $\mathcal{Z}$, rather than its high-resolution version $\mathcal{Y}$. Since MCL is an end-to-end Compressive Learning method \cite{tran2019multilinear}, the values of $\mathbf{\Phi}_k$ which are used to build or configure the CS device are determined via optimization. Thus, the CS component is simulated at the software level using Eq. (\ref{eq2}) during the optimization stage.    
	
	Given the compressed measurements $\mathcal{Z}$, relevant features for the learning task that preserve the tensor structure of the input signal are synthesized by the FS component. Any arbitrary design that preserves the tensor structure can be used for the FS component. For example, in \cite{tran2020multilinear}, the authors use a highly-nonlinear design which consists of multiple convolution layers for the FS component. To isolate the effect of CS and FS component, here we adopt the original formulation in \cite{tran2019multilinear} which mirrors the sensing step in Eq. (\ref{eq2}) by a multilinear transformation to synthesize new features, i.e.:
	\begin{equation}\label{eq3}
	\mathcal{T} = \mathcal{Z} \times_1 \mathbf{\Theta}_1 \times \dots \times_K \mathbf{\Theta}_K
	\end{equation} 
	where $\mathcal{T} \in \mathbb{R}^{\tilde{I}_1 \times \dots \times \tilde{I}_K}$ denotes the synthesized features and  $\mathbf{\Theta}_k \in \mathbb{R}^{\tilde{I}_k \times M_k}$, $k=1, \dots, K$ denote the parameters of the FS component. 
	
	Finally, the task-specific neural network $\mathfrak{N}$ takes the synthesized feature $\mathcal{T}$ as input and outputs the predicted label. 
	
	Similar to other CL methods \cite{adler2016compressed, lohit2016direct, zisselman2018compressed}, during system optimization, MCL utilizes high-resolution signal $\mathcal{Y}$ (often obtained from standard sensors with higher computational cost than the CS sensor) and the corresponding class label to optimize the system's parameters. That is, the parameters of the three components of the MCL model are jointly optimized to maximize the learning performance using stochastic gradient descend. An important processing step in this process is the MCL model's initialization. In \cite{tran2019multilinear}, the authors propose an initialization scheme that preserves the energy of the signal in the compressed measurements $\mathcal{Z}$. This is done by decomposing $\mathcal{Y}$ using the HOSVD \cite{de2000multilinear}: 
	\begin{equation}\label{eq4}
	\mathcal{Y} = \mathcal{S} \times_1 \mathbf{U}_1 \times \dots \times_K \mathbf{U}_K
	\end{equation}
	where $\mathcal{S} \in \mathbb{R}^{M_1 \times \dots \times M_K}$ and $\mathbf{U}_k \in \mathbb{R}^{M_k \times I_k}$, $k=1, \dots, K$. Then, the CS components are initialized with $\mathbf{\Phi}_k = \mathbf{U}_k^T$. Furthermore, the parameters of the FS component is initialized with values that optimally reconstruct (in the least-square sense) the high-resolution signal $\mathcal{Y}$. This is done by setting the dimensions of the synthesized features $\mathcal{T}$ equal those of high-resolution signal $\mathcal{Y}$, i.e., $\tilde{I}_k = I_k, {}\forall k=1, \dots, K$, and setting $\mathbf{\Theta}_k = \mathbf{U}_k$. 
	
	\section{Method}
	When building a MCL model for deployment, the computational requirements determine the range of feasible dimensions for $\mathcal{Y}$ and $\mathcal{Z}$. That is:
	
	\begin{equation}\label{eq5}
	\begin{aligned}
	& I_k^{\textrm{min}} \leq \; I_k \leq I_k^{\textrm{max}} \\
	& M_k^{\textrm{min}} \leq \; M_k \leq M_k^{\textrm{max}} \\
	& \; \; \forall k=1, \dots, K
	\end{aligned}
	\end{equation}
	where $I_k^{\textrm{min}}$ and $I_k^{\textrm{max}}$ denote the lower- and upper-bounds of the feasible values for $I_k$. 
	
	When $K$, $(I_k^{\textrm{max}} - I_k^{\textrm{min}})$ or $(M_k^{\textrm{max}} - M_k^{\textrm{min}})$ are large, the number of possible combinations of $I_k$ and $M_k$ can be enormous. Thus, the motivation of our work lies in the attempt to efficiently determine an optimal configuration of CS device (i.e., the choice of $I_k$ and $M_k$), without the need of conducting the entire optimization process of MCL for every feasible combination of $I_k$ and $M_k$. One might guess that the higher the resolution $I_k$ and number of compressed measurements $M_k$ are, the higher the learning performance will be. However, this is not necessarily true as it will shown in the Experiment Section of this paper. 
	
	One approach to tackle our problem is to empirically characterize the learning performance in terms of $I_k$ and $M_k$ and seek to find an indicator that reflects the performance ranking. Given a learning problem expressed via the training set, the performance of a MCL model depends on its architectural design. There are three main factors that affect the model's complexity, and thus its learning capacity: the CS device configuration, the FS configuration and the architecture of the task-specific neural network $\mathfrak{N}$. The CS device configuration refers to the resolution of the sensor ($I_1 \times \dots \times I_K$) and the dimensions of the compressed measurements ($M_1 \times \dots \times M_k$), while the FS configuration refers to the dimensions of the synthesized features ($\tilde{I}_1 \times \dots \times \tilde{I}_K$). 
	
	In order to analyze and characterize the learning performance in terms of the CS device configuration, it is important to ultimately limit variations in the FS component and the architecture of $\mathfrak{N}$ when evaluating multiple choices of CS configuration across multiple learning problems. To do so, we fix the architecture of $\mathfrak{N}$ given any configuration of CS component. In addition, we also fix the dimensions of $\mathcal{T}$ to $I_1^{\textrm{max}} \times \dots \times I_K^{\textrm{max}}$ for any given value of $I_k$ and $M_k$. That is, the parameters of the FS component have the following dimensions:
	\begin{equation}\label{eq6}
	\begin{aligned}
	\mathbf{\Theta}_1 \in \; \mathbb{R}^{I_1^{\textrm{max}} \times M_1}, &\; \forall M_1 \in  [M_1^{\textrm{min}}, M_1^{\textrm{max}} ]  \\
	& \vdots \\
	\mathbf{\Theta}_K \in \; \mathbb{R}^{I_K^{\textrm{max}} \times M_K}, &\; \forall M_K \in  [M_K^{\textrm{min}}, M_K^{\textrm{max}} ] 
	\end{aligned}
	\end{equation} 
	
	Since we fix the dimensions of $\mathcal{T}$, we can no longer initialize parameters of the FS component using HOSVD if the resolution of CS component is different from the highest feasible resolution, i.e., $I_1 \times \dots \times I_K \neq I_1^{\textrm{max}} \times \dots \times I_K^{\textrm{max}}$. As it has been shown in \cite{tran2019multilinear}, initialization is a crucial step when optimizing MCL models. Thus, to circumvent the inability to use HOSVD, we propose to use a different initialization strategy that still pertains to preserving energy in $\mathcal{Z}$ and $\mathcal{T}$. 
	
	As mentioned previously in Section II, in order to train any end-to-end CL model, high-resolution signals and the corresponding labels are needed. In our work, we only need to acquire the set of training signals with labels at the highest resolution, i.e., $I_1^{\textrm{max}} \times \dots \times I_K^{\textrm{max}}$ using a standard signal acquisition device with higher computational and time complexity than a CS device. For using training data at a lower resolution, instead of using a different device to acquire at a lower resolution, we simulate them by applying down-sampling to the high-resolution signal $\mathcal{Y} \in \mathbb{R}^{I_1^{\textrm{max}} \times \dots \times I_K^{\textrm{max}}}$. 
	
	Let $\mathbf{S}@(I_1 \times \dots \times I_K) = \{(\mathcal{Y}_i@(I_1 \times \dots \times I_K), c_i) | i=1, \dots, N\}$ denote the training set of $N$ samples at resolution $I_1 \times \dots \times I_K$. $\mathbf{S}@(I_1 \times \dots \times I_K)$ represents the training data that is used to optimize an MCL model with the CS device sampling at resolution $I_1 \times \dots \times I_K$. In order to initialize the parameters of the CS and FS components, we obtain the initial values of $\mathbf{\Phi}_k$ and $\mathbf{\Theta}_k$ ($k=1, \dots, K$) by solving the following optimization problem:
	\begin{equation}\label{eq7}
	\begin{split}
	\argmin_{\{\mathbf{\Phi}_k\}, \{\mathbf{\Theta}_k\}} \sum_{i=1}^{N} & \| \textrm{FS}\big(\textrm{CS}(\mathcal{Y}@(I_1\times \dots \times I_K))\big) \\
	& - \; \mathcal{Y}@(I_1^{\textrm{max}} \times \dots \times I_K^{\textrm{max}}) \|_F^2
	\end{split}
	\end{equation}
	where $\textrm{FS}\big(\textrm{CS}(\mathcal{Y}@(I_1\times \dots \times I_K))\big)$ denotes the features synthesized by the FS component, given the CS device operating at resolution $I_1 \times \dots \times I_K$. In addition, $\|.\|_F$ denotes the Frobenius norm. 
	
	The objective in Eq. (\ref{eq7}) is used to initialize $\mathbf{\Phi}_k$ and $\mathbf{\Theta}_k$ with values that produce features resembling (in the least-square sense) the input signals at the highest resolution. This initialization strategy of the CS and FS components thus resembles the one in \cite{tran2019multilinear}, which uses HOSVD. 
	
	To initialize the parameters of the task-specific neural network $\mathfrak{N}$, we optimize the following objective:
	
	\begin{equation}\label{eq8}
	\argmin_{\mathbf{\Omega}} \sum_{i=1}^{N} \mathsf{L}(\mathfrak{N}(\mathcal{Y}@(I_1^{\textrm{max}} \times \dots \times I_K^{\textrm{max}})); c_i)
	\end{equation}
	where $\mathbf{\Omega}$ denotes the parameters of $\mathfrak{N}$, and $\mathsf{L}$ denotes the inference loss function while $\mathfrak{N}(\mathcal{Y}@(I_1^{\textrm{max}} \times \dots \times I_K^{\textrm{max}})$ denotes the prediction generated by $\mathfrak{N}$ given the high-resolution input $\mathcal{Y}$. 
	
	After applying the initialization steps in Eq. (\ref{eq7}) and Eq. (\ref{eq8}), all parameters of the MCL model are jointly optimized to minimize the inference loss:
	\begin{equation}\label{eq9}
	\argmin_{\{\mathbf{\Phi}_k\}, \{\mathbf{\Theta}_k\}, \mathbf{\Omega}} \sum_{i=1}^{N} \mathsf{L} \Big( \mathfrak{N} \big(\textrm{FS}(\textrm{CS}(\mathcal{Y}@(I_1\times \dots \times I_K)))\big), c_i\Big)
	\end{equation}
	
	We optimize the objective functions in Eqs. (\ref{eq7}), (\ref{eq8}), and (\ref{eq9}) using stochastic gradient descend. In the next section, we provide detailed description of our experimental setup as well as our analysis of the effects of CS device configuration based on the empirical results. 
	
	\section{Experiments}
	
	\begin{table*}
		\begin{center}
			\caption{Test Performances of PubFig83 Dataset. The upper section shows test accuracy while the lower section shows Mean Squared Error (MSE) measured on test set when optimizing Eq. (\ref{eq7}). \textbf{Bold-face} numbers indicate the top-3 accuracy and the corresponding MSE}\label{t1}
			\resizebox{0.9\textwidth}{!}{
				\begin{tabular}{|c|c|c|c|c|c|c|}
					\hline
					\multicolumn{2}{|c|}{\multirow{2}{*}{Test Accuracy (\%)}} & \multicolumn{5}{c|}{$\mathcal{Y}$ Dimension $(I_1 \times I_2 \times I_3)$}                                                         \\ \cline{3-7} 
					\multicolumn{2}{|c|}{}                          & $256\times 256\times 3$      & $224\times 224\times 3$     & $192\times 192\times 3$  & $160\times 160\times 3$    & $128\times 128\times 3$      \\ \hline
					\multirow{6}{*}{\makecell{$\mathcal{Z}$ \\ Dimension \\ $(M_1\times M_2\times M_3)$}}      & $30\times 30\times 1$     & $66.01$ & $74.38$ & $67.69$ & $78.36$ & $44.49$ \\ \cline{2-7} 
					& $28\times 28\times 1$    & $\mathbf{80.86}$ & $\mathbf{79.46}$ & $57.17$ & $72.12$ & $58.17$ \\ \cline{2-7} 
					& $26\times 26\times 1$     & $77.53$ & $\mathbf{79.32}$ & $67.47$ & $47.03$ & $47.59$ \\ \cline{2-7} 
					& $24\times 24\times 1$     & $58.54$ & $53.70$ & $54.43$ & $62.00$ & $58.85$ \\ \cline{2-7} 
					& $22\times 22\times 1$     & $57.53$ & $71.16$ & $72.23$ & $77.39$ & $75.67$ \\ \cline{2-7} 
					& $20\times 20\times 1$     & $58.44$ & $68.17$ & $38.35$ & $43.23$ & $71.45$ \\ \hline \hline
					
					\multicolumn{2}{|c|}{\multirow{2}{*}{MSE during initialization (Eq. (\ref{eq7}))}} & \multicolumn{5}{c|}{$\mathcal{Y}$ Dimension $(I_1 \times I_2 \times I_3)$}                                                         \\ \cline{3-7} 
					\multicolumn{2}{|c|}{}                          & $256\times 256\times 3$      & $224\times 224\times 3$     & $192\times 192\times 3$  & $160\times 160\times 3$    & $128\times 128\times 3$      \\ \hline
					\mr{6}{*}{\makecell{$\mathcal{Z}$ \\ Dimension \\ $(M_1\times M_2\times M_3)$}}      & $30\times 30\times 1$     & $0.1368$ & $0.1161$ & $0.0408$ & $0.0256$ & $0.2490$ \\ \cline{2-7} 
					& $28\times 28\times 1$     & $\mathbf{0.0185}$ & $\mathbf{0.0196}$ & $0.0395$ & $0.1357$ & $0.1957$ \\ \cline{2-7} 
					& $26\times 26\times 1$     & $0.0192$ & $\mathbf{0.0173}$ & $0.0320$ & $0.2008$ & $0.2135$ \\ \cline{2-7} 
					& $24\times 24\times 1$     & $0.0480$ & $0.1078$ & $0.1675$ & $0.0567$ & $0.0425$ \\ \cline{2-7} 
					& $22\times 22\times 1$     & $0.1927$ & $0.0241$ & $0.0306$ & $0.0167$ & $0.0189$ \\ \cline{2-7} 
					& $20\times 20\times 1$     & $0.0323$ & $0.0254$ & $0.1218$ & $0.0616$ & $0.0199$ \\ \hline
				\end{tabular}
			}
		\end{center}
	\end{table*}
	
	\begin{figure*}
		\centering
		\includegraphics[width=0.9\textwidth]{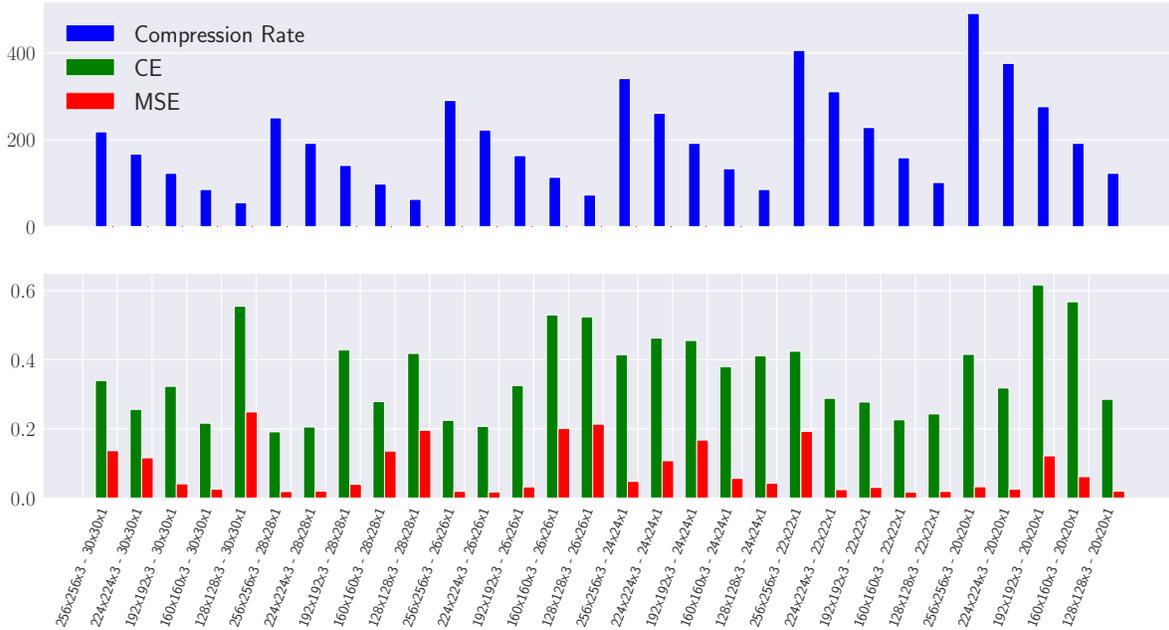}
		\caption{PubFig83 Performance}\label{figure1}
	\end{figure*}

	\begin{table*}
		\begin{center}
			\caption{Test Performances of Caltech101 Dataset. The upper section shows test accuracy while the lower section shows Mean Squared Error (MSE) measured on test set when optimizing Eq. (\ref{eq7}). \textbf{Bold-face} numbers indicate the top-3 accuracy and the corresponding MSE}\label{t2}
			\resizebox{0.9\textwidth}{!}{
				\begin{tabular}{|c|c|c|c|c|c|c|}
					\hline
					\multicolumn{2}{|c|}{\multirow{2}{*}{Test Accuracy (\%)}} & \multicolumn{5}{c|}{$\mathcal{Y}$ Dimension $(I_1 \times I_2 \times I_3)$}                                                         \\ \cline{3-7} 
					\multicolumn{2}{|c|}{}                          & $256\times 256\times 3$      & $224\times 224\times 3$     & $192\times 192\times 3$  & $160\times 160\times 3$    & $128\times 128\times 3$      \\ \hline
					\multirow{6}{*}{\makecell{$\mathcal{Z}$ \\ Dimension \\ $(M_1\times M_2\times M_3)$}}      & $30\times 30\times 1$     
					& $53.35$ & $\mathbf{71.08}$ & $\mathbf{71.16}$ & $61.67$ & $53.67$ \\ \cline{2-7} 
					
					& $28\times 28\times 1$    & $60.68$ & $47.48$ & $65.18$ & $60.09$ & $64.22$ \\ \cline{2-7} 
					& $26\times 26\times 1$     & $54.88$ & $48.15$ & $47.75$ & $57.22$ & $58.02$ \\ \cline{2-7} 
					& $24\times 24\times 1$     & $\mathbf{68.72}$ & $54.77$ & $52.17$ & $55.79$ & $63.46$ \\ \cline{2-7} 
					& $22\times 22\times 1$     & $60.78$ & $53.03$ & $56.87$ & $61.48$ & $54.32$ \\ \cline{2-7} 
					& $20\times 20\times 1$     & $47.72$ & $52.90$ & $53.00$ & $51.42$ & $47.34$ \\ \hline \hline
					
					\multicolumn{2}{|c|}{\multirow{2}{*}{MSE during initialization (Eq. (\ref{eq7}))}} & \multicolumn{5}{c|}{$\mathcal{Y}$ Dimension $(I_1 \times I_2 \times I_3)$}                                                         \\ \cline{3-7} 
					\multicolumn{2}{|c|}{}                          & $256\times 256\times 3$      & $224\times 224\times 3$     & $192\times 192\times 3$  & $160\times 160\times 3$    & $128\times 128\times 3$      \\ \hline
					\mr{6}{*}{\makecell{$\mathcal{Z}$ \\ Dimension \\ $(M_1\times M_2\times M_3)$}}      & $30\times 30\times 1$     
					& $0.3617$ & $\mathbf{0.0692}$ & $\mathbf{0.0396}$ & $0.2611$ & $0.2655$ \\ \cline{2-7} 
					& $28\times 28\times 1$     & $0.1045$ & $0.3819$ & $0.0569$ & $0.2000$ & $0.0638$ \\ \cline{2-7} 
					& $26\times 26\times 1$     & $0.1243$ & $0.3693$ & $0.3571$ & $0.2018$ & $0.2488$ \\ \cline{2-7} 
					& $24\times 24\times 1$     & $\mathbf{0.0837}$ & $0.1753$ & $0.3200$ & $0.1513$ & $0.1823$ \\ \cline{2-7} 
					& $22\times 22\times 1$     & $0.1035$ & $0.2127$ & $0.1805$ & $0.1694$ & $0.3151$ \\ \cline{2-7} 
					& $20\times 20\times 1$     & $0.3857$ & $0.2061$ & $0.2322$ & $0.1522$ & $0.3746$ \\ \hline
				\end{tabular}
			}
		\end{center}
	\end{table*}

	\subsection{Datasets and Experiment Protocol}
	We conducted our empirical analysis using image data. Two image datasets representing two different learning tasks were used in our experiments: face recognition and object recognition. These datasets are:
	\begin{itemize}
		\item PubFig83 \cite{pinto2011scaling} is a medium-size dataset that contains $13002$ facial images of $83$ public figures. The dataset was curated from the list of URLs compiled by \cite{kumar2009attribute} by removing near-duplicate samples and individuals with few samples. Since the photos were collected from the internet, the dataset represents the task of recognizing identities in uncontrolled situations using near-frontal faces. 
		
		\item Caltech101 \cite{fei2004learning} is an object recognition dataset that contains pictures of objects from $101$ categories. Besides $101$ categories, the dataset also contains a background class which represents non-object images. The dataset is not well-balanced with the number of images per category ranging from $40$ to $800$. In total, there are $9145$ images in this dataset. 
	\end{itemize}
	
	For both datasets, we randomly selected $60\%$, $20\%$, $20\%$ of the samples from each class for training, validation and testing, respectively. PubFig83 and Caltech101 both contain RGB images of varying resolutions. In order to simulate different resolutions of the CS device, we resized the images to $5$ different resolutions, ranging from $256 \times 256 \times 3$ to $128 \times 128 \times 3$. That is, we experimented with the set of feasible resolutions of $\mathcal{Y}$: $I_1 \times I_2 \times I_3 \in \{256 \times 256 \times 3, 224 \times 224 \times 3, 192 \times 192 \times 3, 160 \times 160 \times 3, 128 \times 128 \times 3 \}$. Regarding compressed measurements $\mathcal{Z}$, we considered the following set of $6$ feasible dimensions: $M_1 \times M_2 \times M_3 \in \{30 \times 30 \times 1, 28 \times 28 \times 1, 26 \times 26 \times 1, 24 \times 24 \times 1, 22 \times 22 \times 1, 20\times 20\times 1 \}$. This leads to $30$ combinations for the sizes of $\mathcal{Y}$ and $\mathcal{Z}$. For each combination, the experiment was run $5$ times and the average performance on the test set is reported. 
	
	Regarding the architecture of the task-specific neural network $\mathfrak{N}$, we adopted the DenseNet121 architecture proposed in \cite{huang2017densely}, which was pretrained on the ILSVRC2012 database. We first performed the initialization of $\mathfrak{N}$ by optimizing Eq. (\ref{eq8}) for each dataset. To limit the possible variations in the effect of $\mathfrak{N}$ to different CS configurations, the values of $\mathbf{\Omega}$ obtained by optimizing Eq. (\ref{eq8}) is used in all experiments and all combinations of $I_1 \times I_2 \times I_3$ and $M_1 \times M_2 \times M_3$. Different from $\mathfrak{N}$, the initialization of CS and FS components using Eq. (\ref{eq7}) is repeated for every experiment. 
	
	Stochastic optimization was done using ADAM optimizer \cite{kingma2014adam}. Eq. (\ref{eq7}) was optimized for a total of $35$ epochs with the learning rate schedule $\{10^{-3}, 10^{-4}, 10^{-5}\}$, changing at epoch $6$ and $26$. In addition, weight decay regularization of $5\times 10^{-5}$ was used when optimizing Eq. (\ref{eq7}). For optimizing Eqs. (\ref{eq8}) and (\ref{eq9}), we updated the parameters for $120$ epochs starting with learning rate of $10^{-3}$, then dropping to $10^{-4}$, and to $10^{-5}$, at epoch $21$ and $101$, respectively. The weight decay coefficient was set to $10^{-4}$.

	\begin{figure*}
		\centering
		\includegraphics[width=0.9\textwidth]{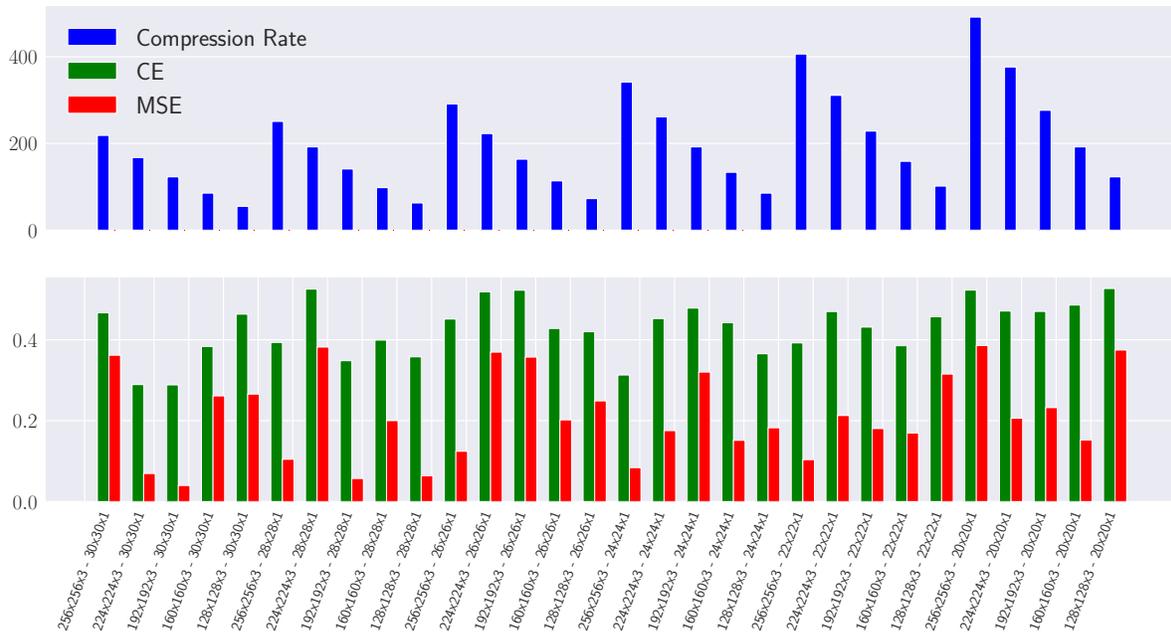}
		\caption{Caltech101 Performance}\label{figure2}
	\end{figure*}
	
	\subsection{Experiment Results}
	
	The test accuracies obtained by using different configurations on PubFig83 and Caltech101 datasets are illustrated in the upper section of Tables \ref{t1} and \ref{t2}, respectively. Moreover, in the lower section of Tables \ref{t1} and \ref{t2}, we also show the Mean Squared Error (MSE) measured on the test set obtained when optimizing Eq. (\ref{eq7}), i.e., \textit{during initializing the CS and FS components}. 
	
	The first observation from our experimental results is that higher resolutions of the CS device and higher numbers of measurements do not always yield better learning performance. In fact, for both datasets, at the maximum resolution ($256 \times 256\times 3$) and the maximum number of compressed measurements ($30\times 30\times 1$), we obtain test accuracies that are far below the best achieved and highlighted with bold-face numbers. 
	
	On a closer look, no clear monotonic relationship between the learning performance and the CS resolution or the number of measurements can be observed from both datasets. For example, when we fix the number of measurements and increase or decrease the CS resolution, we do not observe the corresponding increase or decrease in test accuracy. Similarly, when we fix the CS resolution, the learning performances do not change linearly with the number of measurements. 
	
	On the other hand, the MSE obtained during the initialization of the CS and FS components reflects well the final learning performances. For example, by inspecting the top-3 configurations for both datasets, we can see that the corresponding MSE values are among the lowest. Similarly, those configurations with high MSE values achieve very poor accuracies. 
	
	To better illustrate the trend, we plot the classification error (CE) versus MSE as well as the compression rate ($(I_1 * I_2 * I_3) / (M_1 * M_2 * M_3)$) for PubFig83 and Caltech101 in Figures \ref{figure1} and \ref{figure2}, respectively. By observing both figures, it can be seen that the compression rate shows no clear linear relationship with the learning performance. Quantitatively, the Pearson correlation values between the final classification error (CE) and the MSE during initialization are equal to $\mathbf{0.65}$ and $\mathbf{0.82}$ for PubFig83 and Caltech101, indicating a strong correlation between the final performance and the performance obtained when initializing CS and FS components. On the other hand, the Pearson correlation values between CE and the compression rate are equal to $\mathbf{-0.02}$ and $\mathbf{0.23}$ for PubFig83 and Caltech101, respectively.

	\section{Conclusion}
	Multilinear Compressive Learning (MCL) is an efficient framework to tackle the problem of learning with compressed measurements from high-dimensional multidimensional signals. In this paper, we empirically investigated the learning performance of Multilinear Compressive Learning models with respect to the configurations of the Compressive Sensing device in MCL. Our analysis showed that higher sensor resolutions and higher number of measurements do not always lead to better learning performance. In addition, the compression rate also showed no clear linear relationship with the final learning performance. On the other hand, the Mean Squared Error (MSE) obtained during initializing the CS and FS components of MCL strongly correlates with the final learning performance. This suggests that this metric can be used as a surrogate measure of the final learning performance to gauge between different configurations of the CS device without conducting the entire optimization procedure, which is often time-consuming.

	\section{Acknowledgement}
	This project has received funding from the European Union's Horizon 2020 research and innovation programme under grant agreement No 871449 (OpenDR). This publication reflects the authors’ views only. The European Commission is not responsible for any use that may be made of the information it contains.
	
	\bibliography{reference}
	\bibliographystyle{ieeetr}

\end{document}